\PassOptionsToPackage{prologue,dvipsnames}{xcolor}

\documentclass{article} 
\usepackage{iclr2025_conference,times}

\usepackage{amsmath,amsfonts,bm}

\def\eqref#1{equation~\ref{#1}}

\def\1{\bm{1}}

\DeclareMathAlphabet{\mathsfit}{\encodingdefault}{\sfdefault}{m}{sl}
\SetMathAlphabet{\mathsfit}{bold}{\encodingdefault}{\sfdefault}{bx}{n}

\usepackage{url}
\usepackage[switch]{lineno}
\usepackage{array}
\usepackage{stfloats}
\usepackage{verbatim}
\usepackage{cite}
\usepackage{amsmath,amssymb,amsfonts}
\usepackage{algorithmic}
\usepackage{graphicx}
\usepackage{subfigure}
\usepackage{textcomp}
\usepackage{colortbl}
\usepackage{float}
\usepackage{booktabs}
\definecolor{awesome}{rgb}{0.98, 0.66, 0.68}
\usepackage{multirow}
\usepackage{etoolbox}
\usepackage{setspace}
\usepackage{etoolbox}
\usepackage{algorithm}
\usepackage{pifont}
\usepackage[dvipsnames]{xcolor}
\usepackage[colorlinks=true, linkcolor=blue, citecolor=blue, urlcolor=magenta]{hyperref}
\usepackage{balance}

\newcommand{\cmark}{\textcolor{ForestGreen}{\ding{51}}}  
\newcommand{\xmark}{\textcolor{Red}{\ding{55}}}         

\title{Learning Novel Skills from Language-Generated Demonstrations}

\author{Ao-Qun Jin, Tian-Yu Xiang
\thanks{Ao-Qun Jin is also with the School of Computer and Information Engineering, Hubei Normal University, Huangshi 435002, China. 
Tian-Yu Xiang is also with the School of Artificial Intelligence, University of Chinese Academy of Sciences, Beijing 100049, China.
These authors contributed equally to this work.
} \\
Institute of Automation \\
Chinese Academy of Sciences \\
Beijing 100190, China \\
\And
Xiao-Hu Zhou
\thanks{Corresponding author: Xiao-Hu Zhou.} \\
Institute of Automation \\
Chinese Academy of Sciences \\
Beijing 100190, China \\
\AND
Mei-Jiang Gui, Xiao-Liang Xie, Shi-Qi Liu, Shuang-Yi Wang \\
Institute of Automation \\
Chinese Academy of Sciences \\
Beijing 100190, China \\
\AND
Yue Cao, Sheng-Bin Duan, Fu-Chao Xie, Zeng-Guang Hou
\thanks{Sheng-Bin Duan is also with the School of Information, Shanxi University of Finance and Economics, Taiyuan, 030006, China.} \\
Institute of Automation \\
Chinese Academy of Sciences \\
Beijing 100190, China \\
}

\iclrfinalcopy 
\begin{document}

\maketitle

\begin{abstract}
Robots are increasingly deployed across diverse domains to tackle tasks requiring novel skills. However, current robot learning algorithms for acquiring novel skills often rely on demonstration datasets or environment interactions, resulting in high labor costs and potential safety risks. To address these challenges, this study proposes DemoGen, a skill-learning framework that enables robots to acquire novel skills from natural language instructions. DemoGen leverages the vision-language model and the video diffusion model to generate demonstration videos of novel skills, which enabling robots to learn new skills effectively. Experimental evaluations in the MetaWorld simulation environments demonstrate the pipeline's capability to generate high-fidelity and reliable demonstrations. Using the generated demonstrations, various skill learning algorithms achieve an accomplishment rate three times the original on novel tasks. These results highlight a novel approach to robot learning, offering a foundation for the intuitive and intelligent acquisition of novel robotic skills. (Project website: \href{https://aoqunjin.github.io/LNSLGD/}{DemoGen})
\end{abstract}

\section{Introduction}

\begin{figure}[ht]
    \centering
    \includegraphics[width=\columnwidth]{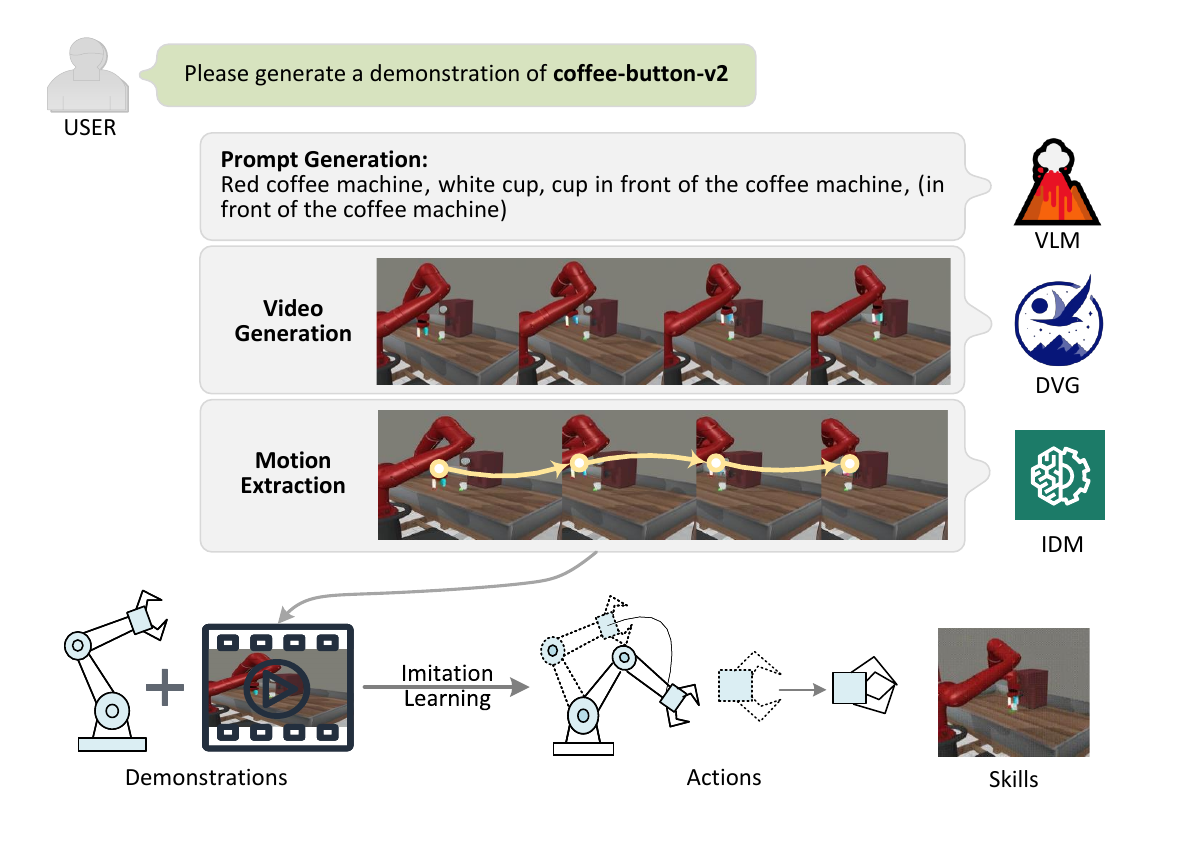} 
    \caption{Demonstration of the novel skill learning steps of the DemoGen's pipeline. For each task, VLM generates an extended text description. With the extended text description, a DVG generates the demonstration videos. Finally, these videos undergo an inverse dynamic model IDM to extract action labels. Robots can learn from generated demonstrations and acquire novel tasks.}
    \label{fig_1}
\end{figure}

Robots have been deployed across various domains, including home assistance, healthcare, and industrial automation~\citep{argall2009survey, rozo2016learning}. In these dynamic environments, they are often required to perform diverse novel tasks that demand novel skills. 
Novel tasks are defined as previously unseen or unfamiliar challenges that demand innovative solutions. Successfully addressing these tasks necessitates the development of novel skills—capabilities that go beyond prior experience or learned behaviors.
Current approaches for skill learning, such as imitation learning (IL) and reinforcement learning (RL), rely on demonstration datasets or environment interactions. These limitations make autonomous novel skill learning in robotics still a challenging and unresolved problem.

RL enables robots to acquire novel skills through environmental interaction and can be applied across environments. However, RL faces significant challenges due to its reliance on trial-and-error exploration, which is time-consuming and potentially dangerous. In both simulated and real-world settings~\citep{yu2020meta, kalashnikov2018scalable}, RL requires extensive interactions between robots and their environments, demanding significant computational resources and time. Additionally, during the RL training phase, robots may exhibit uncontrolled behaviors, posing safety risks in sensitive domains such as home assistance and healthcare~\citep{pecka2014safe, garcia2015comprehensive, gu2022review}. These limitations make RL resource-intensive and challenging to scale for acquiring a wide range of novel skills.

IL enbles robots acquire novel skills through learning the mapping between actions and states in expert demonstrations, bypassing the challenges of environment exploration~\citep{shridhar2023perceiver, reed2022generalist, brohan2023rt, mandlekar2018roboturk}. However, IL relies on the availability of high-quality expert demonstrations, which are often difficult and costly to obtain, particularly in real-world environments~\citep{walke2023bridgedata, o2023open}. Collecting such demonstrations requires skilled operators to carefully configure environments and perform tasks, making the process labor-intensive. Moreover, the unique settings for each novel task in simulated and real-world environments necessitate additional effort, further complicating data collection. In summary, IL is constrained by extensive human involvement, significantly limiting its ability to learn novel skills.

Generative models, including large language models (LLMs)~\citep{mann2020language, touvron2023llama, chowdhery2023palm}, image and video diffusion models\citep{rombach2022high, liu2024sora}, have recently emerged as promising tools for assiting robots to acquire novel skills. These models demonstrate remarkable performance in generating diverse and creative content across text, images, and videos from textual instructions. Trained on massive datasets, they capture underlying patterns, facts, and physical laws, allowing them to generalize and produce accurate results even in unseen scenarios. This capability has inspired their application in robotics, where generating diverse behaviors and solving complex tasks require a deep understanding of task dynamics.

In robotics, generative models such as vision-language models (VLM), vision-language-action models (VLAs), and diffusion models, have demonstrated significant potential across diverse robotic tasks such as planning, generating action instructions, and executing complex behaviors. VLMs~\citep{brohan2023can, driess2023palm} leverage the prior knowledge of large language models to generate action instructions for sub-tasks. Extending this, VLAs incorporate action tokens~\citep{brohan2022rt, brohan2023rt}, enabling the direct output of control signals based on language model priors. Beyond language-based planning, diffusion models—both image- and video-based—predict future frames for task planning and introduce task executors conditioned on visual inputs~\citep{black2023zero, du2024learning}. Despite their impressive performance, these approaches are limited by their reliance on explicit re-training or additional data collection to acquire novel skills autonomously.

This study proposes DemoGen, a skill-learning framework that enables robots to acquire novel skills directly from natural language instructions by leveraging the capabilities of generative models, inverse dynamics model (IDM) and imitation learning model (ILM). The framework utilizes generative models' prior knowledge to generate diverse demonstration videos from textual task descriptions. These demonstrations are then processed through a IDM to construct state-action pairs for ILM's imitation learning, without requiring additional data collection. By leveraging natural language task descriptions, the proposed pipeline allows robots to learn new skills (see Fig~\ref{fig_1}). The primary contributions of this work are summarized as follows:

\begin{itemize}
    \item A learning framework that enables robots to acquire novel skills leveraging generative models, ILM and IDM techniques.
    \item A demonstration generation method is proposed to produce diverse and accurate task demonstrations using the vision language model and the video diffusion model.
    \item Experimental results validate the effectiveness of the DemoGen's pipeline, demonstrating its ability to generate accurate and diverse task demonstrations. On novel tasks, DemoGen achieve an accomplishment rate approximately three times higher than the original.
\end{itemize}

\begin{figure}[!t]
   \centering
     \includegraphics[width=\columnwidth]{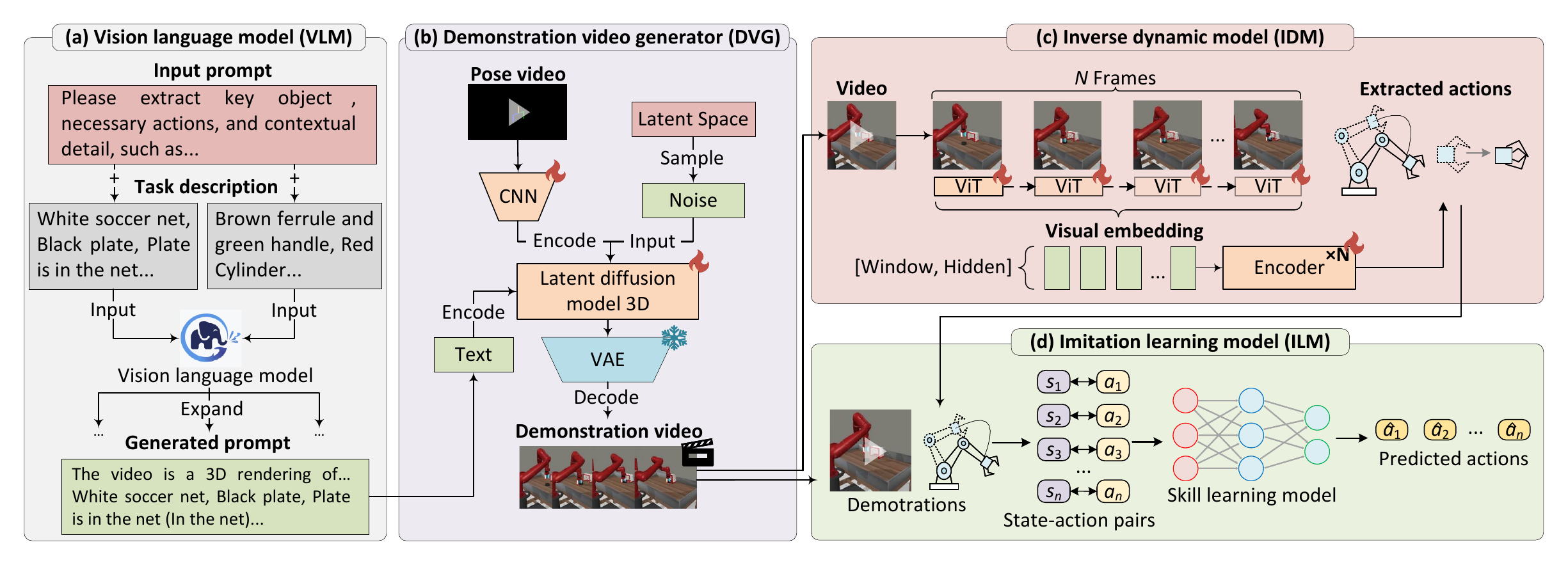}
       \caption{An overview of DemoGen. The task learning process involves four modules: vision language model (a), demonstration video generator (b), inverse dynamic model (c) and imitation learning model (d).}
    \centering
    \label{fig_2}
\end{figure}

\section{Method}

DemoGen comprises four modules that enable robots to learn novel tasks directly from language descriptions (see Fig~\ref{fig_2}). First, a large vision-language model (VLM) enriches task descriptions. These enhanced descriptions are input to a demonstration video generator (DVG), which synthesizes task-specific video demonstrations. Next, an inverse dynamics model (IDM) extracts paired actions and states from these demonstrations. Finally, an end-to-end imitation learning model (ILM) maps environmental states to actions, enabling robots to acquire novel skills.

\subsection{Demonstration Generation}

\subsubsection{Prompt Expansion with VLM}

VLM is employed to expand concise task descriptions~\citep{glm2024chatglm}. Compared to the single modal LLM (with only language as input), VLMs, trained with both language and visual modalities, offer a deeper understanding of real-world contexts. The prompt expansion process can be formalized as follows:

\begin{equation}
P(Y|X) = \prod_{t=1}^T P(Y_t|Y_{<t}, X)
\end{equation}
where $X$ represents the input context, $Y$ denotes the generated prompt sequence, and $Y_{<t}$ indicates the previously generated tokens. This formulation describes the autoregressive nature of the language generation process, where each token is conditioned on the input and previously generated sequence.

\begin{table}[t]
\caption{Examples of prompt expansion by VLM}
\label{table_1}
\begin{center}
\renewcommand{\arraystretch}{1.2}
\begin{tabular}{|m{3.5cm}|m{5.5cm}|}
\hline
\multicolumn{1}{|c|}{\bf Task descriptions} & \multicolumn{1}{c|}{\bf Extended prompts} \\
\hline
Vertical button, reddish-brown wall & Vertical button, reddish-brown wall, (Wall) \\
\hline
Brown ferrule and green handle, red cylinder, rings over columns & Brown ferrule and green handle, red cylinder, rings over columns, (Over columns) \\
\hline
Black metal safety door, door close, door lock & Black metal safety door, door close, door lock, (Close), (Lock) \\
\hline
\end{tabular}
\end{center}
\end{table}

The few-shot learning strategy is employed in VLM to expand prompt descriptions~\citep{mann2020language} without additional training. During inference, manually written examples guide the model in understanding the task requirements. Relevant keywords, such as key objects, required actions, and environmental context (see Table~\ref{table_1}), are first extracted. The VLM emphasizes these essential keywords, ensuring that the generated prompts and demonstrations are precise and aligned with the task objectives.

\subsubsection{Demonstration Video Generation with DVG}

With the enriched prompts generated by the VLM, a diffusion-based text-to-video model serves as the demonstration video generator (DVG). The DVG requires only a limited amount of video data for fine-tuning. It employs a two-stage diffusion process—forward diffusion and reverse diffusion, as described in IDDPM~\citep{nichol2021improved}—to capture intricate task details. The objective is to generate high-quality demonstration videos that visually represent the described tasks, including robot movements, object interactions, and environmental details.

The process of DVG training is illustrated in Algorithm~\ref{alg_1}. Given a video sequence $ V_0 = { v_1^0, v_2^0, \dots, v_t^0 } $, where $ t $ represents the number of frames and $ v_t^0 $ denotes the $t$-th frame, the forward diffusion process progressively adds noise at each time step:

\begin{equation}
q(V_i | V_{i-1}) = \mathcal{N} \left[ V_i ; \sqrt{\alpha_i} V_{i-1}, (1 - \alpha_i) \mathbf{I} \right]
\end{equation}
where $i$ is the time step, $ \alpha_i $ is a time-dependent noise scaling factor, $ \mathcal{N} $ is a multivariate Gaussian distribution, and $ \mathbf{I} $ is the identity matrix. The sequence transitions to a noisier state with each time step.

The reverse diffusion process aims to reconstruct the original video sequence $ V_0 $ from these noisy frames.

\begin{equation}
p_{\theta_1}(V_{i-1} | V_{i}) = \mathcal{N} \left[ V_{i-1}; \mu_{\theta_1} (V_i, i), \Sigma_{\theta_1} (V_i, i) \right]
\end{equation}
where $ \mu_{\theta_1} $ and $ \Sigma_{\theta_1} $ are the predicted mean and covariance of the noisy frame $ V_i $.

The model is optimized using the following loss function:

\begin{equation}
\mathcal{L}_1 = \mathbb{E}_{i, V_0, \boldsymbol{\epsilon}} \left[ \|\boldsymbol{\epsilon} - \boldsymbol{\epsilon}_{\theta_1}(V_t, i)\|^2 \right]
\end{equation}
where $ \boldsymbol{\epsilon}_{\theta_1} $ is the noise predicted by the model, and $ \boldsymbol{\epsilon} $ is the actual noise added during the forward process. 

Unlike static image diffusion models, the DVG utilizes spatio-temporal attention to capture complex dependencies across both spatial and temporal dimensions, enabling the generation of coherent and realistic video sequences.

The DVG in the proposed framework is fine-tuned using a diverse dataset of robot manipulation videos annotated by the VLM, allowing the model to learn the mapping between textual descriptions and corresponding video content. During the fine-tuning process, the DVG incorporates a T2I-Adapter~\citep{mou2023t2i} to assist the robot arm in executing precise movements. The T2I-Adapter takes pose videos as input, where these pose videos are generated through pose rendering during fine-tuning based on 3D poses from the training data. In addition, the movement of the robot arm is planned by the VLM during the inference step based on the script~\citep{liang2023code}, and the pose videos are rendered based on the script.

\begin{algorithm}[tb]
    \caption{: Training scheme for DVG}
    \begin{algorithmic}[1]
    \REQUIRE $\mathbf{V}$, $\mathbf{X}$, $\mathbf{Y}$, $\mathbf{P}$, $\mathbf{N}$, $B$, $E$ (video, text, generated text, pose, noise video, batch size, epoch number.)
        \STATE Initialize parameters for DVG, pretrained VLM, and noise generator IDDPM.
        \FOR{$i=1$ to $E$}
                \REPEAT
                \STATE Sample $B$ video-text pairs from the training data.
                \STATE \#\# Generate extended text descriptions using VLM.
                \STATE $\mathbf{Y} = \text{VLM}(\mathbf{X})$
                \STATE \#\# Add noise to the video data.
                \STATE $\mathbf{N} = \text{IDDPM}(\mathbf{V})$
                \STATE \#\# Input noise, text and pose to the model.
                \STATE $\hat{\mathbf{V}} = \text{DVG}(\mathbf{N}, \mathbf{Y}, \mathbf{P})$
                \STATE Calculate loss $\mathcal{L}_1$.
                \STATE Update parameters of DVG model based on $\mathcal{L}_1$.
                \UNTIL{all video-text pairs are enumerated.}
        \ENDFOR
    \end{algorithmic}
    \label{alg_1}
\end{algorithm}

\subsubsection{Motion Extraction with IDM}

The IDM extracts actions from demonstration sequences, constructing state-action pairs for imitation learning. From a sequence of states \( S = \{s_1, s_2, \dots, s_n\} \), derived from the demonstration video \( V \) and segmented using a sliding window of size \( n+1 \), the IDM infers the corresponding sequence of actions \( A = \{a_1, a_2, \dots, a_n\} \). The IDM is trained with parameters \( \theta_2 \) to minimize the negative log-likelihood of the true motion intentions conditioned on the observed states. The loss function is defined as:

\begin{equation}
\mathcal{L}_2 = -\sum_{t=1}^{n} \log P_{\theta_2}(a_t | s_{t}, s_{t+1}, \dots, s_{t+n})
\end{equation}
where \( a_t \) denotes the action at time step \( t \), and \( s_{t+n} \) represents the state at time step \( t+n \).

The IDM architecture combines a transformer encoder and a vision transformer (ViT) to capture both temporal and spatial features from the input data. The transformer encoder models temporal dependencies within sequential states by leveraging future observations to predict current motion intentions. Simultaneously, the ViT processes visual data from video frames, learning spatial features critical for identifying patterns in robotic manipulation tasks.

\subsection{Skill Learning}

Once the state-action pairs are obtained, the final step involves establishing a mapping between states and actions through end-to-end imitation learning. This process can be framed as a markov decision process (MDP), which includes elements such as the set of possible states, the set of possible actions, and a transition model that predicts how the system moves between states based on actions. While traditional reinforcement learning relies on a reward function to guide the agent’s behavior, imitation learning focuses on mimicking expert demonstrations without explicitly using rewards. The agent learns a policy by minimizing the difference between its predicted actions and those shown in the expert data. The policy is trained to minimize the behavioral cloning loss:

\begin{equation}
\mathcal{L}_{3} = \mathbb{E}_{(s,a)\sim \mathcal{D}}[\|\pi(s) - a\|^2]
\end{equation}
where $\mathcal{D}$ represents the demonstration dataset, $s$ denotes the state, $a$ is the expert action, and $\pi(s)$ is the action predicted by the policy network.

Policy learning is achieved through a learning objective that reduces the error between the agent’s actions and the expert’s, effectively teaching the agent to replicate the expert’s behavior as closely as possible.

\begin{algorithm}[tb]
    \caption{: Novel skill scheme of the DemoGen's pipeline}
    \begin{algorithmic}[1]
    \REQUIRE $\mathbf{V}$, $\mathbf{X}$, $\mathbf{Y}$, $\mathbf{P}$, $\mathbf{N}$, $M$ (video, task description text, generated text, pose, noise video, number of demonstrations per task.)
        \STATE Initialize parameters for pretrained VLM, DVG, IDM, and noise generator IDDPM.
        \FOR{$i=1$ to $M$} 
            \STATE \#\# Generate extended text description.
            \STATE $\mathbf{Y} = \text{VLM}(\mathbf{X})$
            \STATE \#\# Generate pose sequence.
            \STATE $\hat{\mathbf{P}} = \text{VLM}(\mathbf{Y})$
            
            \STATE \#\# Sample random noise for the video generation.
            \STATE $\mathbf{N} = \text{IDDPM}()$
            
            \STATE \#\# Generate demonstration video using DVG.
            \STATE $\hat{\mathbf{V}} = \text{DVG}(\mathbf{N}, \mathbf{Y}, \hat{\mathbf{P}})$
            
        \ENDFOR
    \end{algorithmic}
    \label{alg_2}
\end{algorithm}

\subsection{Pipeline for Learning Novel Skills}

DemoGen enables zero-shot task learning by using prompts to generate demonstrations for previously unseen tasks, allowing robots to acquire novel skills without prior examples. The process is detailed in Algorithm~\ref{alg_2}. Upon receiving a prompt for a novel task, the pipeline employs the VLM to expand the task description, providing details about the environment and required actions. Based on the enriched description, the DVG generates demonstration videos. The IDM extracts motion intentions from these videos, constructing state-action pairs. Finally, an imitation learning technique maps states to actions, achieving novel skill learning.

\section{Experiments}

\subsection{Data Collection}

To validate the efficiency of the DemoGen's pipeline, 20 manipulation trials are collected for each of the 22 tasks from the MetaWorld~\citep{yu2020meta} multi-task learning benchmark. Data are collected using a fixed camera setup with an elevation angle of $-25^\circ$ and an azimuth of $145^\circ$. The collection frequency is set at 80 Hz, with trajectories capped at a maximum length of 500 steps. Visual frames are rendered at a resolution of $512 \times 512$. To enrich the diversity of state-action pairs, random action selection is incorporated during the data collection with an epsilon-greedy probability of 0.1.

\subsection{Training Details}

\subsubsection{Experimental Setup}

The 22 tasks are organized into 16 groups, which are further divided into two folds to define few-shot and zero-shot tasks. Performance evaluation of the DemoGen's pipeline is conducted under the two-fold cross-validation (task compositions are detailed in Table~\ref{table_2}).

\begin{table}[!t]
\caption{Two-Fold cross-validation based on set A and set B}
\label{table_2}
\begin{center}
\renewcommand{\arraystretch}{1.2}
\begin{tabular}{|p{5cm}<{\centering}|p{5cm}<{\centering}|}
\hline
\multicolumn{1}{|c|}{\bf Fold 1} & \multicolumn{1}{c|}{\bf Fold 2} \\
\hline
button (topdown, wall) & button (topdown-wall, press) \\
\hline
reach & reach (wall) \\
\hline
push (wall) & push \\
\hline
door (close, lock) & door (open, unlock) \\
\hline
drawer (open) & drawer (close) \\
\hline
faucet (open) & faucet (close) \\
\hline
plate (slide, side-back) & plate (slide-back, side) \\
\hline
window (open) & window (close) \\
\hline
\end{tabular}
\end{center}
\end{table}

\subsubsection{Preprocessing}

During the training stage of the DVG, video data are sampled with a skip step of 3, yielding up to 36 frames per video. In contrast, the IDM and ILM are trained with the original frame rate. The IDM applies edge extraction and noise injection as preprocessing steps, while the ILM uses normalization and Gaussian noise to improve robustness across varying states.

\subsubsection{Model Configuration}

The VLM in the DemoGen's pipeline is the closed-source commercial model GLM-4-0520~\citep{glm2024chatglm}, which supports direct few-shot inference. Three handwritten examples are provided for VLM to expand prompts. The DVG is based on the Tune-A-Video~\citep{wu2023tune}, initialized with Stable-Diffusion-1.4 and enhanced with randomly initialized Spatio-Temporal Attention and a T2I-Adapter. The IDM is based on ViT and incorporates a randomly initialized Transformer-Encoder to encode temporal dimensions.

The DVG is fine-tuned using 20 demonstration videos, expanded task descriptions, and corresponding poses for each few-shot task. Similarly, the IDM is trained with 20 demonstration videos and corresponding actions for each few-shot tasks. Input data for the IDM is segmented with a temporal window size of 12. The learning rate and batch size were optimized using the Ray Tune hyperparameter optimization framework.

Two skill learning models, LCBC~\citep{stepputtis2020language} and RT-1~\citep{brohan2022rt}, are employed as ILM due to their strong generalization and task-learning ability. LCBC utilizes T5~\citep{raffel2020exploring} text embeddings to encode language instructions and Vision Transformer (ViT)~\citep{dosovitskiy2020image} features to extract visual information. These multimodal encodings are processed by a transformer-encoder-based policy network to integrate linguistic and visual data for action prediction. RT-1 extends this approach by incorporating historical states and using a transformer-decoder with self-attention to process temporal sequences of visual and language features, enhancing action prediction across multiple timesteps.

The task learning baselines are trained on 20 demonstration samples for each few-shot task, with an additional 11 zero-shot tasks used to test zero-shot performance and use a discrete action space of 256 bins. In contrast, the DemoGen's pipeline uses the generated demonstrations, covering both few-shot and zero-shot tasks. The learning rate and batch size were similarly optimized using Ray Tune.

\begin{figure}[!t]
   \centering
     \includegraphics[width=\columnwidth]{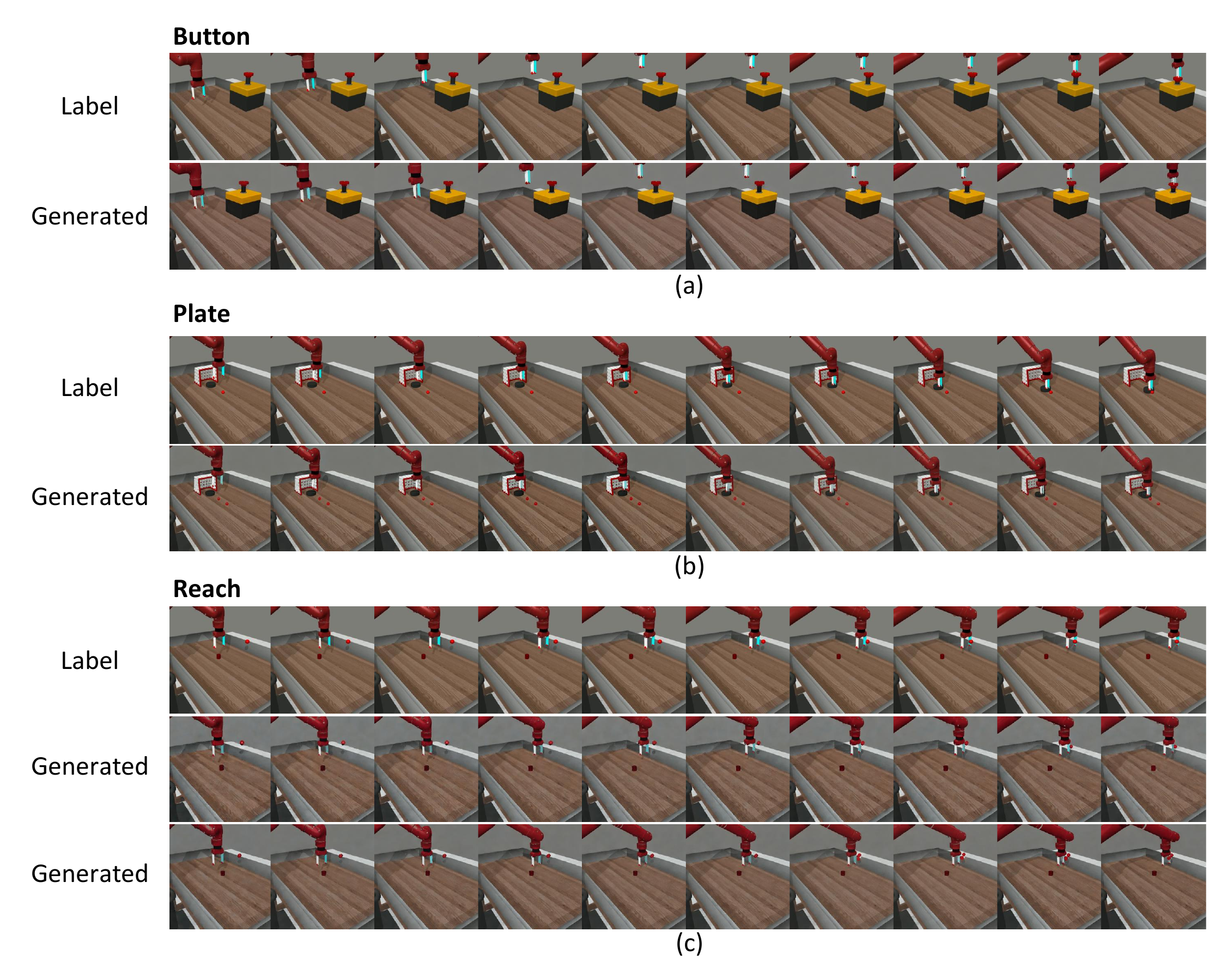}
       \caption{The proposed framework can generate demonstrations that show fidelity, diversity (a, b) and creativity (c).}
    \centering
    \label{fig_3}
\end{figure}

\subsection{Experimental Metrics}

\subsubsection{Metrics for Demonstation Generation}

The quality of generated demonstrations is a critical factor in the DemoGen's pipeline. Human evaluators are recruited to rate each generated demonstration based on the following criteria:

\begin{itemize} 
    \item \textbf{Physical laws:} Adherence to physical laws and principles.
    \item \textbf{Accomplishment:} Successful completion of the presented task.
    \item \textbf{Consistency:} Consistency between the video content and its description. 
\end{itemize}

\subsubsection{Metrics for Skill Learning}

The proposed framework aims to enable robots to learn novel skills, with the primary metric being task accomplishment. This study reports task-level metrics for each fold and the overall accomplishment rate across tasks at the fold level.

\subsection{Experimental Results}

The experimental results are reported under two setups:
\begin{itemize} 
\item \textbf{Few-Shot Learning:} The DemoGen's pipeline is fine-tuned using a small number of manipulation trials, representing few-shot skill learning. In this setup, results are reported on the same tasks used during the fine-tuning stage. 
\item \textbf{Zero-Shot Learning:} The DemoGen's pipeline is evaluated on tasks not seen during fine-tuning, directly testing its ability to learn novel tasks. 
\end{itemize}

\subsubsection{Quality of the Generated Demonstration}

\begin{table}[t]
\caption{Human evaluation results comparing Few-Shot Tasks, Zero-Shot Tasks, and overall performance across three criteria: adherence to physical laws, task accomplishment, and consistency with descriptions. Tuning Tasks consistently achieved high scores, particularly in physical, while Zero-Shot Tasks most times performed well, especially in description consistency.}
\label{table_3}
\begin{center}
\renewcommand{\arraystretch}{1.2}
\begin{tabular}{|l|c|c|c|}
\hline
\textbf{Method} & \textbf{Physical Laws} & \textbf{Task Accomplishment} & \textbf{Consistency} \\
\hline
Few-Shot Learning & 88.3\% & 92.5\% & 96.1\% \\
\hline
Zero-Shot Learning & 57.9\% & 63.2\% & 71.6\% \\
\hline
{Average} & {73.1\%} & {77.8\%} & {83.8\%} \\
\hline
\end{tabular}
\end{center}
\end{table}

The quality of the generated demonstrations is evaluated by four human evaluators, with average scores across three metrics reported under two experimental settings in Fig~\ref{fig_3}. In both settings, the generated demonstrations adhered to physical laws, successfully accomplished tasks, and aligned with descriptions in over $50\%$ of cases. Although performance slightly decreased under the few-shot learning setting, the above-$50\%$ success rate demonstrates the pipeline's ability to generate reliable demonstrations, contributing to novel skill learning. Visualization results in Table~\ref{table_3} illustrate that the generated demonstrations achieve both fidelity and diversity.

\subsubsection{Skill Learning}

Under the few-shot learning setting, different imitation algorithms achieved comparable performance using the generated demonstrations from DemoGen and the collected expert manipulations. This indicates that the quality of the generated demonstrations is similar to the collected manipulations and that the state-action pairs constructed by the IDM are precise enough for the imitation learning pipeline.

Notably, under the zero-shot learning setting, the generated demonstrations achieved nearly three times the task accomplishment rates of algorithms trained with fine-tuned expert-collected data (see Table~\ref{table_4}). This result underscores the DemoGen's superior capability to enable robots to learn novel, unseen tasks, which is challenging for the previous imitation learning methods. Moreover, LCBC demonstrated comparable performance under the zero-shot learning setting to that achieved in the few-shot setting, further confirming the effectiveness of DemoGen in enabling robots to acquire novel skills.

\begin{table*}[!t]\small
\centering
\renewcommand{\arraystretch}{1.2}
\caption{Comparison between learning from expert data and generated data.}
\vspace{1em}
\scalebox{0.75}{
\begin{tabular}{|p{1.3cm}<{\centering}|p{1.3cm}<{\centering}|p{1.3cm}<{\centering}|p{1cm}<{\centering}|p{1cm}<{\centering}|p{1cm}<{\centering}|p{1cm}<{\centering}|p{1cm}<{\centering}|p{1cm}<{\centering}|p{1cm}<{\centering}|p{1cm}<{\centering}|p{1cm}<{\centering}|}
\hline
\textbf{Type} & \textbf{Method} & \textbf{Data} & \textbf{Button} & \textbf{Reach} & \textbf{Push} & \textbf{Door} & \textbf{Drawer} & \textbf{Faucet} & \textbf{Plate} & \textbf{Window} & \textbf{Total} \\
\hline
\multirow{4}{*}{Few-Shot} & \multirow{2}{*}{LCBC} & Expert & \cmark & \cmark & \xmark & \cmark & \xmark & \xmark & \cmark & \cmark & $5 \ / \ 8$ \\
\cline{3-12}
& & Generated & \cmark & \cmark & \cmark & \cmark & \xmark & \xmark & \cmark & \cmark & \textbf{6} $/$ \textbf{8} \\
\cline{2-12}
& \multirow{2}{*}{RT-1} & Expert & \cmark & \cmark & \cmark & \cmark & \cmark & \cmark & \cmark & \cmark & \textbf{8} $/$ \textbf{8} \\
\cline{3-12}
& & Generated & \xmark & \cmark & \xmark & \cmark & \xmark & \xmark & \cmark & \cmark & $4 \ / \ 8$ \\
\hline
\multirow{4}{*}{Zero-Shot} & \multirow{2}{*}{LCBC} & Expert & \cmark & \xmark & \xmark & \xmark & \cmark & \xmark & \xmark & \xmark & $2 \ / \ 8$ \\
\cline{3-12}
& & Generated & \cmark & \cmark & \xmark & \cmark & \cmark & \cmark & \xmark & \cmark & \textbf{6} $/$ \textbf{8} \\
\cline{2-12}
& \multirow{2}{*}{RT-1} & Expert & \cmark & \xmark & \xmark & \xmark & \cmark & \xmark & \xmark & \xmark & $2 \ / \ 8$ \\
\cline{3-12}
& & Generated & \cmark & \cmark & \xmark & \xmark & \cmark & \cmark & \xmark & \cmark & \textbf{5} $/$ \textbf{8} \\
\hline
\end{tabular}}
\label{table_4}
\end{table*}

\section{Conclusions}

This study proposes DemoGen, enabling robots to learn novel skills directly from natural language-based instructions, leveraging generative models, IDM and ILM. Specifically, an generative-models-based method organizes actions by utilizing the prior knowledge embedded in generative models, directly converting language instructions into demonstration videos. IDM and ILM algorithms are then applied to learn novel skills from these generated demonstrations. 
This approach addresses limitations in task data collection, which are constrained by labor-intensive manual construction and physical restrictions. Future work will focus on refining and optimizing the framework to enhance its accuracy and efficiency in robot skill learning, particularly for complex manipulation tasks. Additionally, the proposed pipeline will undergo rigorous validation in real-world environments to assess its scalability and practical applicability.

\subsubsection*{Acknowledgments}

This work was supported in part by the National Key Research and Development Program of China under 2023YFC2415100, in part by the National Natural Science Foundation of China under Grant 62222316, Grant 62373351, Grant 82327801, Grant 62073325, Grant 62303463, in part by the Chinese Academy of Sciences Project for Young Scientists in Basic Research under Grant No.YSBR-104 and in part by China Postdoctoral Science Foundation under Grant 2024M763535.

\bibliography{iclr2025_conference}

\begin{thebibliography}{31}
\providecommand{\natexlab}[1]{#1}
\providecommand{\url}[1]{\texttt{#1}}
\expandafter\ifx\csname urlstyle\endcsname\relax
  \providecommand{\doi}[1]{doi: #1}\else
  \providecommand{\doi}{doi: \begingroup \urlstyle{rm}\Url}\fi

\bibitem[Argall et~al.(2009)Argall, Chernova, Veloso, and Browning]{argall2009survey}
Brenna~D Argall, Sonia Chernova, Manuela Veloso, and Brett Browning.
\newblock A survey of robot learning from demonstration.
\newblock \emph{Robotics and Autonomous Systems}, 57\penalty0 (5):\penalty0 469--483, 2009.

\bibitem[Black et~al.(2023)Black, Nakamoto, Atreya, Walke, Finn, Kumar, and Levine]{black2023zero}
Kevin Black, Mitsuhiko Nakamoto, Pranav Atreya, Homer Walke, Chelsea Finn, Aviral Kumar, and Sergey Levine.
\newblock Zero-shot robotic manipulation with pretrained image-editing diffusion models.
\newblock \emph{arXiv preprint arXiv:2310.10639}, 2023.

\bibitem[Brohan et~al.(2023{\natexlab{a}})Brohan, Chebotar, Finn, Hausman, Herzog, Ho, Ibarz, Irpan, Jang, Julian, et~al.]{brohan2023can}
Anthony Brohan, Yevgen Chebotar, Chelsea Finn, Karol Hausman, Alexander Herzog, Daniel Ho, Julian Ibarz, Alex Irpan, Eric Jang, Ryan Julian, et~al.
\newblock Do as i can, not as i say: Grounding language in robotic affordances.
\newblock In \emph{Conference on Robot Learning}, pp.\  287--318. PMLR, 2023{\natexlab{a}}.

\bibitem[Brohan et~al.(2022)]{brohan2022rt}
Anthony Brohan et~al.
\newblock R{T}-1: Robotics transformer for real-world control at scale.
\newblock \emph{arXiv preprint arXiv:2212.06817}, 2022.

\bibitem[Brohan et~al.(2023{\natexlab{b}})]{brohan2023rt}
Anthony Brohan et~al.
\newblock R{T}-2: Vision-language-action models transfer web knowledge to robotic control.
\newblock \emph{arXiv preprint arXiv:2307.15818}, 2023{\natexlab{b}}.

\bibitem[Chowdhery et~al.(2023)]{chowdhery2023palm}
Aakanksha Chowdhery et~al.
\newblock Palm: Scaling language modeling with pathways.
\newblock \emph{Journal of Machine Learning Research}, 24\penalty0 (240):\penalty0 1--113, 2023.

\bibitem[Dosovitskiy et~al.(2020)]{dosovitskiy2020image}
Alexey Dosovitskiy et~al.
\newblock An image is worth 16x16 words: Transformers for image recognition at scale.
\newblock \emph{arXiv preprint arXiv:2010.11929}, 2020.

\bibitem[Driess et~al.(2023)]{driess2023palm}
Danny Driess et~al.
\newblock Palm-e: An embodied multimodal language model.
\newblock \emph{arXiv preprint arXiv:2303.03378}, 2023.

\bibitem[Du et~al.(2024)Du, Yang, Dai, Dai, Nachum, Tenenbaum, Schuurmans, and Abbeel]{du2024learning}
Yilun Du, Sherry Yang, Bo~Dai, Hanjun Dai, Ofir Nachum, Josh Tenenbaum, Dale Schuurmans, and Pieter Abbeel.
\newblock Learning universal policies via text-guided video generation.
\newblock \emph{Advances in Neural Information Processing Systems}, 36, 2024.

\bibitem[Garc{\i}a \& Fern{\'a}ndez(2015)Garc{\i}a and Fern{\'a}ndez]{garcia2015comprehensive}
Javier Garc{\i}a and Fernando Fern{\'a}ndez.
\newblock A comprehensive survey on safe reinforcement learning.
\newblock \emph{Journal of Machine Learning Research}, 16\penalty0 (1):\penalty0 1437--1480, 2015.

\bibitem[GLM et~al.(2024)GLM, Zeng, Xu, Wang, Zhang, Yin, Zhang, Rojas, Feng, Zhao, et~al.]{glm2024chatglm}
Team GLM, Aohan Zeng, Bin Xu, Bowen Wang, Chenhui Zhang, Da~Yin, Dan Zhang, Diego Rojas, Guanyu Feng, Hanlin Zhao, et~al.
\newblock Chatglm: A family of large language models from glm-130b to glm-4 all tools.
\newblock \emph{arXiv preprint arXiv:2406.12793}, 2024.

\bibitem[Gu et~al.(2024)Gu, Yang, Du, Chen, Walter, Wang, and Knoll]{gu2022review}
Shangding Gu, Long Yang, Yali Du, Guang Chen, Florian Walter, Jun Wang, and Alois Knoll.
\newblock A review of safe reinforcement learning: Methods, theories, and applications.
\newblock \emph{IEEE Transactions on Pattern Analysis and Machine Intelligence}, 46\penalty0 (12):\penalty0 11216--11235, 2024.

\bibitem[Kalashnikov et~al.(2018)]{kalashnikov2018scalable}
Dmitry Kalashnikov et~al.
\newblock Scalable deep reinforcement learning for vision-based robotic manipulation.
\newblock In \emph{Proceedings of the Conference on Robot Learning}, pp.\  651--673, 2018.

\bibitem[Liang et~al.(2023)Liang, Huang, Xia, Xu, Hausman, Ichter, Florence, and Zeng]{liang2023code}
Jacky Liang, Wenlong Huang, Fei Xia, Peng Xu, Karol Hausman, Brian Ichter, Pete Florence, and Andy Zeng.
\newblock Code as policies: Language model programs for embodied control.
\newblock In \emph{Proceedings of the 2023 IEEE International Conference on Robotics and Automation}, pp.\  9493--9500, 2023.

\bibitem[Liu et~al.(2024)]{liu2024sora}
Yixin Liu et~al.
\newblock Sora: A review on background, technology, limitations, and opportunities of large vision models.
\newblock \emph{arXiv preprint arXiv:2402.17177}, 2024.

\bibitem[Mandlekar et~al.(2018)]{mandlekar2018roboturk}
Ajay Mandlekar et~al.
\newblock Roboturk: A crowdsourcing platform for robotic skill learning through imitation.
\newblock In \emph{Proceedings of the Conference on Robot Learning}, pp.\  879--893, 2018.

\bibitem[Mann et~al.(2020)Mann, Ryder, Subbiah, Kaplan, Dhariwal, Neelakantan, Shyam, Sastry, Askell, Agarwal, et~al.]{mann2020language}
Ben Mann, N~Ryder, M~Subbiah, J~Kaplan, P~Dhariwal, A~Neelakantan, P~Shyam, G~Sastry, A~Askell, S~Agarwal, et~al.
\newblock Language models are few-shot learners.
\newblock \emph{arXiv preprint arXiv:2005.14165}, 1, 2020.

\bibitem[Mou et~al.(2023)Mou, Wang, Xie, Wu, Zhang, Qi, Shan, and Qie]{mou2023t2i}
Chong Mou, Xintao Wang, Liangbin Xie, Yanze Wu, Jian Zhang, Zhongang Qi, Ying Shan, and Xiaohu Qie.
\newblock T2i-adapter: Learning adapters to dig out more controllable ability for text-to-image diffusion models.
\newblock \emph{arXiv preprint arXiv:2302.08453}, 2023.

\bibitem[Nichol \& Dhariwal(2021)Nichol and Dhariwal]{nichol2021improved}
Alexander~Quinn Nichol and Prafulla Dhariwal.
\newblock Improved denoising diffusion probabilistic models.
\newblock In \emph{Proceedings of the International Conference on Machine Learning}, pp.\  8162--8171, 2021.

\bibitem[O'Neill et~al.(2023)]{o2023open}
Abby O'Neill et~al.
\newblock Open x-embodiment: Robotic learning datasets and rt-x models.
\newblock \emph{arXiv preprint arXiv:2310.08864}, 2023.

\bibitem[Pecka \& Svoboda(2014)Pecka and Svoboda]{pecka2014safe}
Martin Pecka and Tomas Svoboda.
\newblock Safe exploration techniques for reinforcement learning--an overview.
\newblock In \emph{Proceedings of the Modelling and Simulation for Autonomous Systems: First International Workshop, MESAS 2014, Rome}, pp.\  357--375. Springer, 2014.

\bibitem[Raffel et~al.(2020)Raffel, Shazeer, Roberts, Lee, Narang, Matena, Zhou, Li, and Liu]{raffel2020exploring}
Colin Raffel, Noam Shazeer, Adam Roberts, Katherine Lee, Sharan Narang, Michael Matena, Yanqi Zhou, Wei Li, and Peter~J Liu.
\newblock Exploring the limits of transfer learning with a unified text-to-text transformer.
\newblock \emph{The Journal of Machine Learning Research}, 21\penalty0 (1):\penalty0 5485--5551, 2020.

\bibitem[Reed et~al.(2022)]{reed2022generalist}
Scott Reed et~al.
\newblock A generalist agent.
\newblock \emph{Transactions on Machine Learning Research}, 2022.

\bibitem[Rombach et~al.(2022)Rombach, Blattmann, Lorenz, Esser, and Ommer]{rombach2022high}
Robin Rombach, Andreas Blattmann, Dominik Lorenz, Patrick Esser, and Bj{\"o}rn Ommer.
\newblock High-resolution image synthesis with latent diffusion models.
\newblock In \emph{Proceedings of the IEEE/CVF Conference on Computer Vision and Pattern Recognition}, pp.\  10684--10695, 2022.

\bibitem[Rozo et~al.(2016)Rozo, Calinon, Caldwell, Jimenez, and Torras]{rozo2016learning}
Leonel Rozo, Sylvain Calinon, Darwin~G Caldwell, Pablo Jimenez, and Carme Torras.
\newblock Learning physical collaborative robot behaviors from human demonstrations.
\newblock \emph{IEEE Transactions on Robotics}, 32\penalty0 (3):\penalty0 513--527, 2016.

\bibitem[Shridhar et~al.(2023)Shridhar, Manuelli, and Fox]{shridhar2023perceiver}
Mohit Shridhar, Lucas Manuelli, and Dieter Fox.
\newblock Perceiver-actor: A multi-task transformer for robotic manipulation.
\newblock In \emph{Proceedings of the Conference on Robot Learning}, pp.\  785--799, 2023.

\bibitem[Stepputtis et~al.(2020)Stepputtis, Campbell, Phielipp, Lee, Baral, and Ben~Amor]{stepputtis2020language}
Simon Stepputtis, Joseph Campbell, Mariano Phielipp, Stefan Lee, Chitta Baral, and Heni Ben~Amor.
\newblock Language-conditioned imitation learning for robot manipulation tasks.
\newblock \emph{Advances in Neural Information Processing Systems}, 33:\penalty0 13139--13150, 2020.

\bibitem[Touvron et~al.(2023)]{touvron2023llama}
Hugo Touvron et~al.
\newblock Llama: Open and efficient foundation language models.
\newblock \emph{arXiv preprint arXiv:2302.13971}, 2023.

\bibitem[Walke et~al.(2023)]{walke2023bridgedata}
Homer~Rich Walke et~al.
\newblock Bridgedata v2: A dataset for robot learning at scale.
\newblock In \emph{Proceedings of the Conference on Robot Learning}, pp.\  1723--1736, 2023.

\bibitem[Wu et~al.(2023)Wu, Ge, Wang, Lei, Gu, Shi, Hsu, Shan, Qie, and Shou]{wu2023tune}
Jay~Zhangjie Wu, Yixiao Ge, Xintao Wang, Stan~Weixian Lei, Yuchao Gu, Yufei Shi, Wynne Hsu, Ying Shan, Xiaohu Qie, and Mike~Zheng Shou.
\newblock Tune-a-video: One-shot tuning of image diffusion models for text-to-video generation.
\newblock In \emph{Proceedings of the IEEE/CVF International Conference on Computer Vision}, pp.\  7623--7633, 2023.

\bibitem[Yu et~al.(2020)Yu, Quillen, He, Julian, Hausman, Finn, and Levine]{yu2020meta}
Tianhe Yu, Deirdre Quillen, Zhanpeng He, Ryan Julian, Karol Hausman, Chelsea Finn, and Sergey Levine.
\newblock Meta-world: A benchmark and evaluation for multi-task and meta reinforcement learning.
\newblock In \emph{Proceedings of the Conference on Robot Learning}, pp.\  1094--1100, 2020.

\end{thebibliography}
\bibliographystyle{iclr2025_conference}

\appendix
\section{Appendix}

\subsection{Prompt Expansion Details}

\begin{figure}[!t]
   \centering
     \includegraphics[width=\columnwidth]{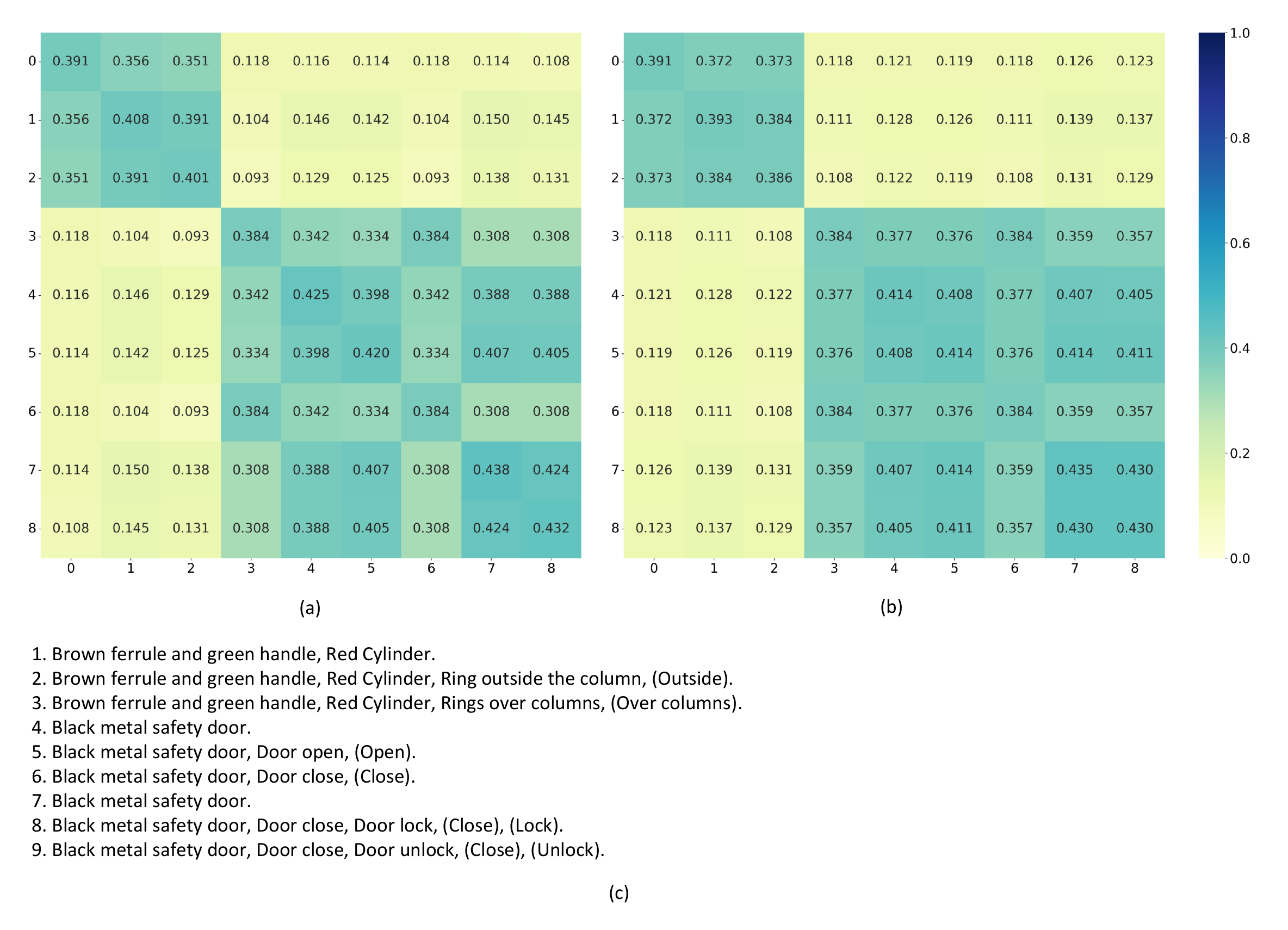}
       \caption{The cosine similarity matrices between prompt embeddings of different tasks (c). The embeddings of expanded prompts (a) exhibit higher cosine similarity scores than the original prompts (b).}
    \centering
    \label{fig_4}
\end{figure}

A comprehensive similarity analysis is conducted to evaluate the effectiveness of the prompt expansion. The analysis utilized cosine similarity metrics to quantify the relationship between different prompt embeddings, as illustrated in Figure~\ref{fig_4}. The results show that prompts associated with semantically similar tasks consistently exhibit higher cosine similarity scores, indicating successful clustering in the embedding space. This clustering behavior suggests that the extended prompts effectively capture task-relevant features and semantic relationships, providing a strong foundation for subsequent demonstration video generation.

\subsection{Demonstration Generation Details}

\begin{figure}[!t]
   \centering
     \includegraphics[width=\columnwidth]{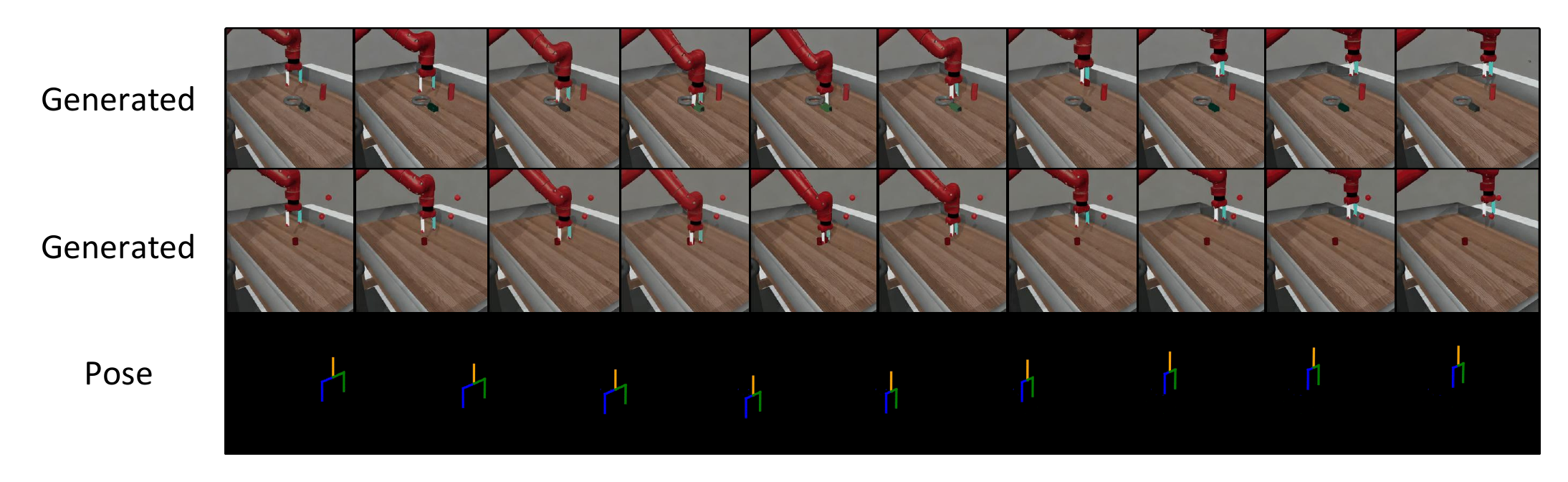}
       \caption{The frame sequences of the generated robot actions, which keep the consistency with the input trajectories.}
    \centering
    \label{fig_5}
\end{figure}

\begin{figure}[!t]
   \centering
     \includegraphics[width=\columnwidth]{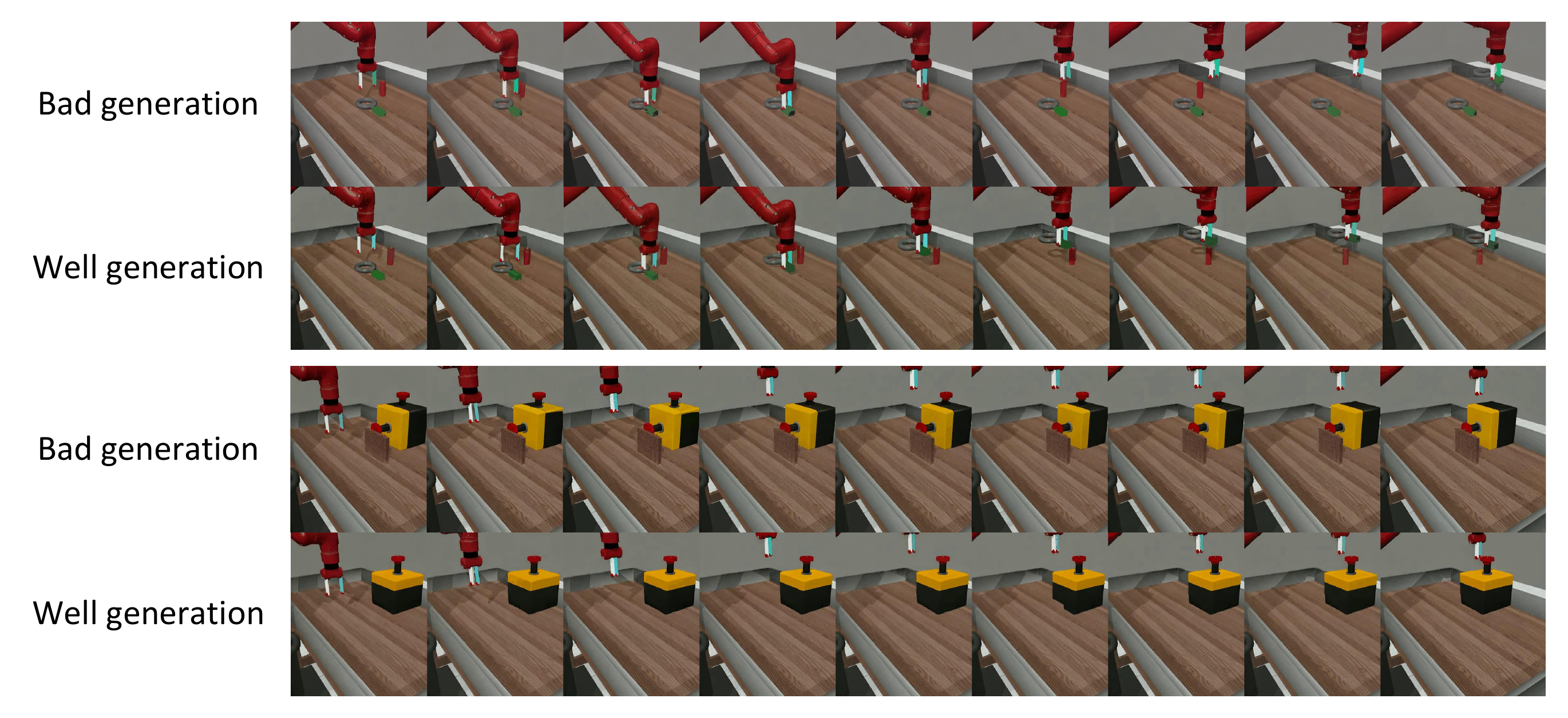}
       \caption{Examples of failure cases in DVG-generated demonstrations.}
    \centering
    \label{fig_6}
\end{figure}

The DVG exhibits robust capabilities in generating consistent robot action sequences based on the input trajectories, as demonstrated in Figure~\ref{fig_5}. The generated demonstrations show high fidelity to the intended actions while maintaining adaptability across diverse scenarios. However, as illustrated in Figure~\ref{fig_6}, the DVG can produce erroneous outputs, particularly when attempting to generate demonstrations for novel tasks that significantly deviate from the training distribution. The experimental results show that these errors occur in approximately 40\% of cases involving novel tasks.
A designed validation module could assess demonstration quality and remove the wrong generations. This would contribute to maintaining reliability of the proposed pipeline.

\subsection{Generalization capabilities of IDM}

\begin{table*}[!htb]\small
\caption{Performance comparison of IDM under different training configurations. The results show the accuracy for both few-shot and zero-shot learning scenarios across varying numbers of tasks and trajectories. Higher percentages indicate better generalization performance.}
\vspace{1em}
\centering
\renewcommand{\arraystretch}{1.2}
\begin{tabular}{|p{3cm}<{\centering}|p{1.5cm}<{\centering}|p{1.5cm}<{\centering}|p{1.5cm}<{\centering}|p{1.5cm}<{\centering}|}
\hline
\multirow{2}{*}{\textbf{Method}} & \multicolumn{2}{c|}{\textbf{5 Tasks}} & \multicolumn{2}{c|}{\textbf{45 Tasks}} \\
\cline{2-5}
& \textbf{50 Traj.} & \textbf{100 Traj.} & \textbf{50 Traj.} & \textbf{100 Traj.} \\
\hline
Few-Shot Learning & \textbf{86.9\%} & 86.1\% & 83.1\% & 84.1\%\\
\hline
Zero-Shot Learning & 77.8\% & 76.0\% & \textbf{81.6\%} & 80.3\% \\
\hline
\end{tabular}
\label{table_5}
\end{table*}

A comprehensive evaluation of the IDM's generalization capabilities was conducted, with results presented in Table~\ref{table_5}. The analysis reveals that task diversity plays a more critical role in generalization performance than the number of trajectories per task. Models trained on fewer than five tasks showed limited generalization capabilities, regardless of the number of trajectories per task. In contrast, expanding the training set to forty-five distinct tasks resulted in substantially improved generalization performance, particularly in zero-shot learning scenarios. 

\end{document}